\documentclass{article}

\PassOptionsToPackage{numbers}{natbib}
\usepackage[final]{nips_2016}
\usepackage[pdftex]{graphicx}
\DeclareGraphicsExtensions{.pdf,.jpeg,.png}

\usepackage[utf8]{inputenc} 
\usepackage[T1]{fontenc}    
\usepackage{url}            
\usepackage{booktabs}       
\usepackage{array}          
\usepackage{amsfonts}       
\usepackage{nicefrac}       
\usepackage{microtype}      
\usepackage{wrapfig}        
\usepackage{subcaption}
\usepackage{adjustbox}
\usepackage{multirow}

\newcolumntype{C}{>{\centering\arraybackslash}m{\dimexpr.1\linewidth-1\tabcolsep}}
\newcolumntype{M}{>{\centering\arraybackslash}m{\dimexpr.225\linewidth-1\tabcolsep}}

\title{ENet: A Deep Neural Network Architecture for Real-Time Semantic Segmentation}

\author{
  Adam Paszke\\
  Faculty of Mathematics, Informatics and Mechanics\\
  University of Warsaw, Poland \\
  \texttt{a.paszke@students.mimuw.edu.pl} \\
  \And
  Abhishek Chaurasia, Sangpil Kim, Eugenio Culurciello\\
  Electrical and Computer Engineering\\
  Purdue University, USA\\
  \texttt{aabhish, sangpilkim, euge@purdue.edu}\\
}

\begin{document}

\maketitle

\begin{abstract}
The ability to perform pixel-wise semantic segmentation in real-time is of paramount importance in mobile applications.
Recent deep neural networks aimed at this task have the disadvantage of requiring a large number of floating point operations and have long run-times that hinder their usability.
In this paper, we propose a novel deep neural network architecture named ENet (efficient neural network), created specifically for tasks requiring low latency operation.
ENet is up to 18$\times$ faster, requires 75$\times$ less FLOPs, has 79$\times$ less parameters, and provides similar or better accuracy to existing models.
We have tested it on CamVid, Cityscapes and SUN datasets and report on comparisons with existing state-of-the-art methods, and the trade-offs between accuracy and processing time of a network.
We present performance measurements of the proposed architecture on embedded systems and suggest possible software improvements that could make ENet even faster.
\end{abstract}

\section{Introduction}

Recent interest in augmented reality wearables, home-automation devices, and self-driving vehicles has created a strong need for semantic-segmentation (or visual scene-understanding) algorithms that can operate in real-time on low-power mobile devices.
These algorithms label each and every pixel in the image with one of the object classes.
In recent years, the availability of larger datasets and computationally-powerful machines have helped deep convolutional neural networks (CNNs) \cite{lecun1998cnn,alex2012,karen14,christian15} surpass the performance of many conventional computer vision algorithms \cite{jamie09,perr2010,vande2011}.
Even though CNNs are increasingly successful at classification and categorization tasks, they provide coarse spatial results when applied to pixel-wise labeling of images.
Therefore, they are often cascaded with other algorithms to refine the results, such as color based segmentation \cite{clement13} or conditional random fields \cite{liang14}, to name a few.

In order to both spatially classify and finely segment images, several neural network architectures have been proposed, such as SegNet \cite{badrinarayanan15basic,badrinarayanan15} or fully convolutional networks \cite{long15}.
All these works are based on a VGG16 \cite{simonyan14} architecture, which is a very large model designed for multi-class classification.
These references propose networks with huge numbers of parameters, and long inference times.
In these conditions, they become unusable for many mobile or battery-powered applications, which require processing images at rates higher than 10 fps.

In this paper, we propose a new neural network architecture optimized for fast inference and high accuracy.
Examples of images segmented using ENet are shown in Figure \ref{fig:intro}.
In our work, we chose not to use any post-processing steps, which can of course be combined with our method, but would worsen the performance of an end-to-end CNN approach.


In Section \ref{architecture} we propose a fast and compact encoder-decoder architecture named ENet.
It has been designed according to rules and ideas that have appeared in the literature recently, all of which we discuss in Section \ref{design}.
Proposed network has been evaluated on Cityscapes \cite{cityscape2016} and CamVid \cite{camvid08} for driving scenario, whereas SUN dataset \cite{sun2015} has been used for testing our network in an indoor situation.
We benchmark it on NVIDIA Jetson TX1 Embedded Systems Module as well as on an NVIDIA Titan X GPU.
The results can be found in Section \ref{results}.

\begin{table}[t]
  \begin{adjustbox}{width=1.0\textwidth,center}
    \small
    \begin{tabular}{CMMM}
      Input image
      &\includegraphics[width=.225\textwidth,trim={1.8cm 0 1.5cm 0},clip]{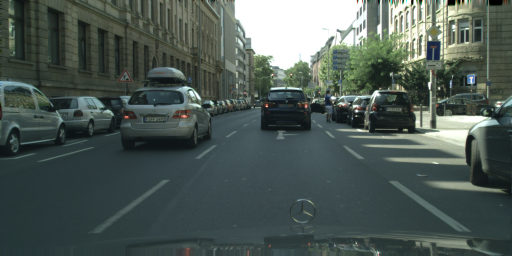}
      &\includegraphics[width=.225\textwidth,trim={0 0 0 2.0cm},clip]{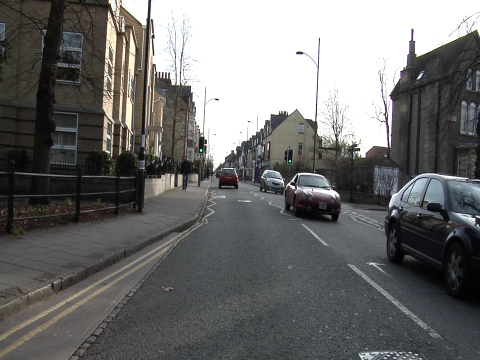}
      &\includegraphics[width=.225\textwidth,trim={0 0 0 1.5cm},clip]{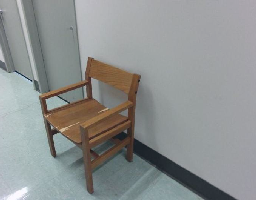}
      \\
      ENet output
      &\includegraphics[width=.225\textwidth,trim={1.8cm 0 1.5cm 0},clip]{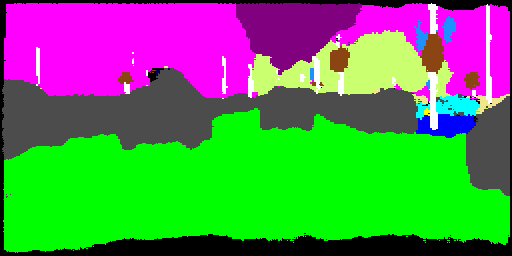}
      &\includegraphics[width=.225\textwidth,trim={0 0 0 2.0cm},clip]{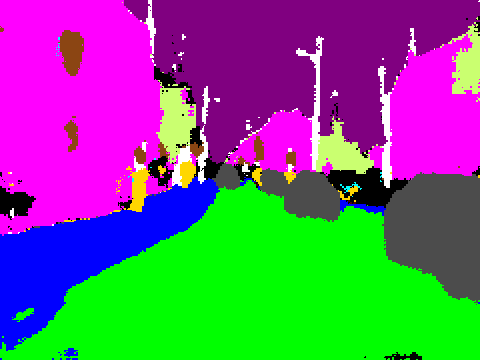}
      &\includegraphics[width=.225\textwidth,trim={0 0 0 1.5cm},clip]{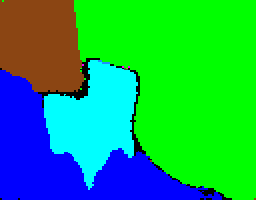}
      \\
    \end{tabular}
  \end{adjustbox}
  \captionof{figure}{ENet predictions on different datasets (left to right Cityscapes, CamVid, and SUN).}
  \label{fig:intro}
\end{table}

\section{Related work}

Semantic segmentation is important in understanding the content of images and finding target objects.
This technique is of utmost importance in applications such as driving aids and augmented reality.
Moreover, real-time operation is a must for them, and therefore, designing CNNs \textit{carefully} is vital.
Contemporary computer vision applications extensively use deep neural networks, which are now one of the most widely used techniques for many different tasks, including semantic segmentation.
This work presents a new neural network architecture, and therefore we aim to compare to other literature that performs the large majority of inference in the same way.

State-of-the-art scene-parsing CNNs use two separate neural network architectures combined together: an encoder and a decoder.
Inspired by probabilistic auto-encoders \cite{ranzato07,ngiam11}, encoder-decoder network architecture has been introduced in SegNet-basic \cite{badrinarayanan15basic}, and further improved in SegNet \cite{badrinarayanan15}.
The encoder is a vanilla CNN (such as VGG16 \cite{simonyan14}) which is trained to classify the input, while the decoder is used to upsample the output of the encoder \cite{long15,noh2015learning,zheng2015conditional,eigen2015predicting,hong2015decoupled}.
However, these networks are slow during inference due to their large architectures and numerous parameters.
Unlike in fully convolutional networks (FCN) \cite{long15}, fully connected layers of VGG16 were discarded in the latest incarnation of SegNet, in order to reduce the number of floating point operations and memory footprint, making it the smallest of these networks.
Still, none of them can operate in real-time.

Other existing architectures use simpler classifiers and then cascade them with Conditional Random Field (CRF) as a post-processing step \cite{liang14,sturgess09}.
As shown in \cite{badrinarayanan15}, these techniques use onerous post-processing steps and often fail to label the classes that occupy fewer number of pixels in a frame.
CNNs can be also combined with recurrent neural networks \cite{zheng2015conditional} to improve accuracy, but then they suffer from speed degradation.
Also, one has to keep in mind that RNN, used as a post-processing step, can be used in conjunction with any other technique, including the one presented in this work.

\section{Network architecture} \label{architecture}

\begin{figure}[!t]
\begin{subfigure}{.5\textwidth}
  \centering
  \includegraphics[width=1.8in]{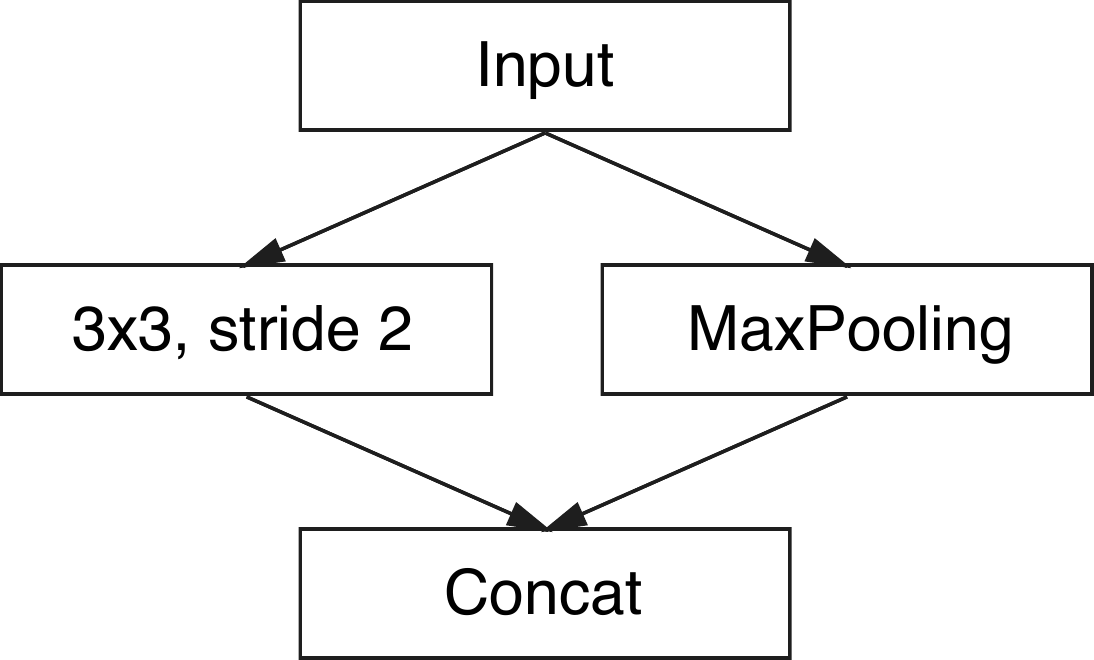}
  \\
  \caption{}
  \label{fig:initial}
\end{subfigure}%
\begin{subfigure}{.5\textwidth}
  \centering
  \includegraphics[height=1.8in]{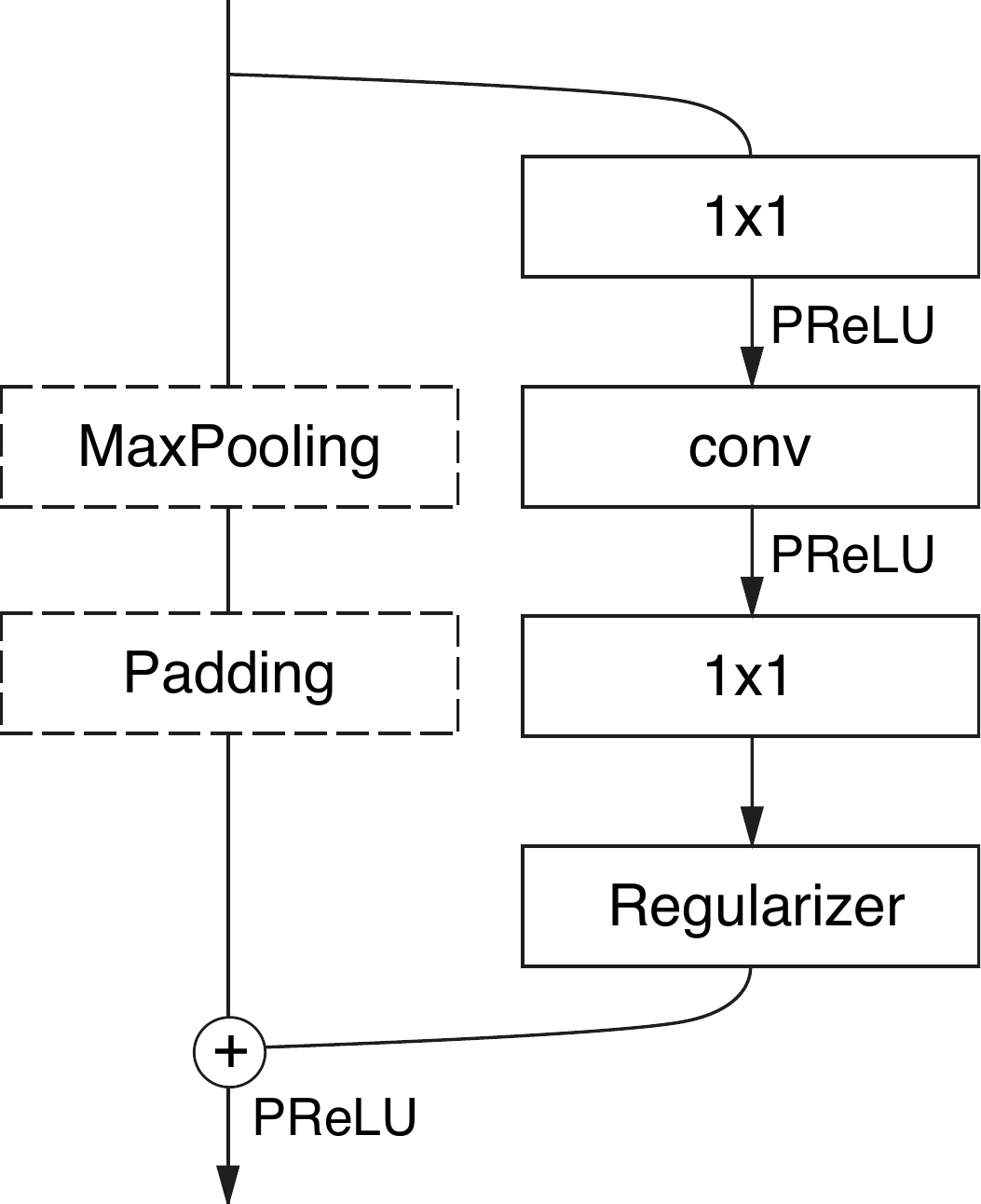}
  \caption{}
  \label{fig:bottleneck}
\end{subfigure}
  \vspace{0.05in}
  \caption{
          (a) ENet initial block. MaxPooling is performed with non-overlapping $2 \times 2$ windows, and the convolution has 13 filters, which sums up to 16 feature maps after concatenation. This is heavily inspired by \cite{szegedy2015rethinking}.
          (b) ENet bottleneck module. \texttt{conv} is either a regular, dilated, or full convolution (also known as deconvolution) with $3 \times 3$ filters, or a $5 \times 5$ convolution decomposed into two asymmetric ones.}
\label{fig:modules}
\end{figure}

The architecture of our network is presented in Table \ref{tab:structure}.
It is divided into several stages, as highlighted by horizontal lines in the table and the first digit after each block name.
Output sizes are reported for an example input image resolution of $512 \times 512$.
We adopt a view of ResNets \cite{he2015resnet} that describes them as having a single main branch and extensions with convolutional filters that separate from it, and then merge back with an element-wise addition, as shown in Figure \ref{fig:bottleneck}.
Each block consists of three convolutional layers: a $1 \times 1$ projection that reduces the dimensionality, a main convolutional layer (\texttt{conv} in Figure \ref{fig:bottleneck}), and a $1 \times 1$ expansion.
We place Batch Normalization \cite{ioffe2015batchnorm} and PReLU \cite{he2015} between all convolutions.
Just as in the original paper, we refer to these as bottleneck modules.
If the bottleneck is downsampling, a max pooling layer is added to the main branch.
\begin{wraptable}{r}{2.75in}
  \footnotesize
  \caption{ENet architecture. Output sizes are given for an example input of $512 \times 512$.}
  \vspace{0.05in}
  \label{tab:structure}
  \centering
  \begin{tabular}{@{}l >{\centering}m{0.7in} c @{}}
    \toprule
    Name                & Type & Output size                  \\
    \midrule
    initial             &                   & $16 \times 256 \times 256$   \\
    \midrule
    bottleneck1.0       &  downsampling     & $64 \times 128 \times 128$   \\
    $4 \times$ bottleneck1.x  &                   & $64 \times 128 \times 128$   \\
    \midrule
    bottleneck2.0       &  downsampling     & $128 \times 64 \times 64$    \\
    bottleneck2.1       &                   & $128 \times 64 \times 64$    \\
    bottleneck2.2       &  dilated $2$      & $128 \times 64 \times 64$    \\
    bottleneck2.3       &  asymmetric $5$   & $128 \times 64 \times 64$    \\
    bottleneck2.4       &  dilated $4$      & $128 \times 64 \times 64$    \\
    bottleneck2.5       &                   & $128 \times 64 \times 64$    \\
    bottleneck2.6       &  dilated $8$      & $128 \times 64 \times 64$    \\
    bottleneck2.7       &  asymmetric $5$   & $128 \times 64 \times 64$    \\
    bottleneck2.8       &  dilated $16$     & $128 \times 64 \times 64$    \\
    \midrule
    \multicolumn{3}{l}{\textit{Repeat section 2, without bottleneck2.0}} \\
    \midrule
    bottleneck4.0       &  upsampling       & $64 \times 128 \times 128$    \\
    bottleneck4.1       &                   & $64 \times 128 \times 128$    \\
    bottleneck4.2       &                   & $64 \times 128 \times 128$    \\
    \midrule
    bottleneck5.0       &  upsampling       & $16 \times 256 \times 256$    \\
    bottleneck5.1       &                   & $16 \times 256 \times 256$    \\
    \midrule
    fullconv            &                   & $C  \times 512 \times 512$    \\
    \bottomrule
  \end{tabular}
  \vspace{-0.2in}
\end{wraptable}
Also, the first $1 \times 1$ projection is replaced with a $2 \times 2$ convolution with stride $2$ in both dimensions.
We zero pad the activations, to match the number of feature maps.
\texttt{conv} is either a regular, dilated or full convolution (also known as deconvolution or fractionally strided convolution) with $3 \times 3$ filters. Sometimes we replace it with asymmetric convolution i.e. a sequence of $5 \times 1$ and $1 \times 5$ convolutions.
For the regularizer, we use Spatial Dropout \cite{tompson15}, with $p = 0.01$ before bottleneck2.0, and $p = 0.1$ afterwards.

The initial stage contains a single block, that is presented in Figure \ref{fig:initial}.
Stage 1 consists of $5$ bottleneck blocks, while stage 2 and 3 have the same structure, with the exception that stage 3 does not downsample the input at the beginning (we omit the $0$th bottleneck). These three first stages are the encoder. Stage 4 and 5 belong to the decoder.

We did not use bias terms in any of the projections, in order to reduce the number of kernel calls and overall memory operations, as cuDNN \cite{chetlur2014cudnn} uses separate kernels for convolution and bias addition.
This choice didn't have any impact on the accuracy.
Between each convolutional layer and following non-linearity we use Batch Normalization \cite{ioffe2015batchnorm}.
In the decoder max pooling is replaced with max unpooling, and padding is replaced with spatial convolution without bias.
We did not use pooling indices in the last upsampling module, because the initial block operated on the $3$ channels of the input frame, while the final output has $C$ feature maps (the number of object classes).
Also, for performance reasons, we decided to place only a bare full convolution as the last module of the network, which alone takes up a sizeable portion of the decoder processing time.

\section{Design choices} \label{design}
In this section we will discuss our most important experimental results and intuitions, that have shaped the final architecture of ENet.

\paragraph{Feature map resolution}
Downsampling images during semantic segmentation has two main drawbacks.
Firstly, reducing feature map resolution implies loss of spatial information like exact edge shape.
Secondly, full pixel segmentation requires that the output has the same resolution as the input.
This implies that strong downsampling will require equally strong upsampling, which increases model size and computational cost.
The first issue has been addressed in FCN \cite{long15} by adding the feature maps produced by encoder, and in SegNet \cite{badrinarayanan15basic} by saving indices of elements chosen in max pooling layers, and using them to produce sparse upsampled maps in the decoder.
We followed the SegNet approach, because it allows to reduce memory requirements.
Still, we have found that strong downsampling hurts the accuracy, and tried to limit it as much as possible.

However, downsampling has one big advantage.
Filters operating on downsampled images have a bigger receptive field, that allows them to gather more context.
This is especially important when trying to differentiate between classes like, for example, rider and pedestrian in a road scene.
It is not enough that the network learns how people look, the context in which they appear is equally important.
In the end, we have found that it is better to use dilated convolutions for this purpose \cite{yu2015dilated}.

\paragraph{Early downsampling}
One crucial intuition to achieving good performance and real-time operation is realizing that processing large input frames is very expensive.
This might sound very obvious, however many popular architectures do not to pay much attention to optimization of early stages of the network, which are often the most expensive by far.

ENet first two blocks heavily reduce the input size, and use only a small set of feature maps.
The idea behind it, is that visual information is highly spatially redundant, and thus can be compressed into a more efficient representation.
Also, our intuition is that the initial network layers should not directly contribute to classification.
Instead, they should rather act as good feature extractors and only preprocess the input for later portions of the network.
This insight worked well in our experiments; increasing the number of feature maps from $16$ to $32$ did not improve accuracy on Cityscapes \cite{cityscape2016} dataset.

\paragraph{Decoder size}
In this work we would like to provide a different view on encoder-decoder architectures than the one presented in \cite{badrinarayanan15}.
SegNet is a very symmetric architecture, as the encoder is an exact mirror of the encoder.
Instead, our architecture consists of a large encoder, and a small decoder.
This is motivated by the idea that the encoder should be able to work in a similar fashion to original classification architectures, i.e. to operate on smaller resolution data and provide for information processing and filtering.
Instead, the role of the the decoder, is to upsample the output of the encoder, only fine-tuning the details.

\paragraph{Nonlinear operations}
A recent paper \cite{he2016identity} reports that it is beneficial to use ReLU and Batch Normalization layers before convolutions.
We tried applying these ideas to ENet, but this had a detrimental effect on accuracy.
Instead, we have found that removing most ReLUs in the initial layers of the network improved the results.
It was quite a surprising finding so we decided to investigate its cause.

\begin{figure}[!t]
  \centering
  \includegraphics[height=1.75in]{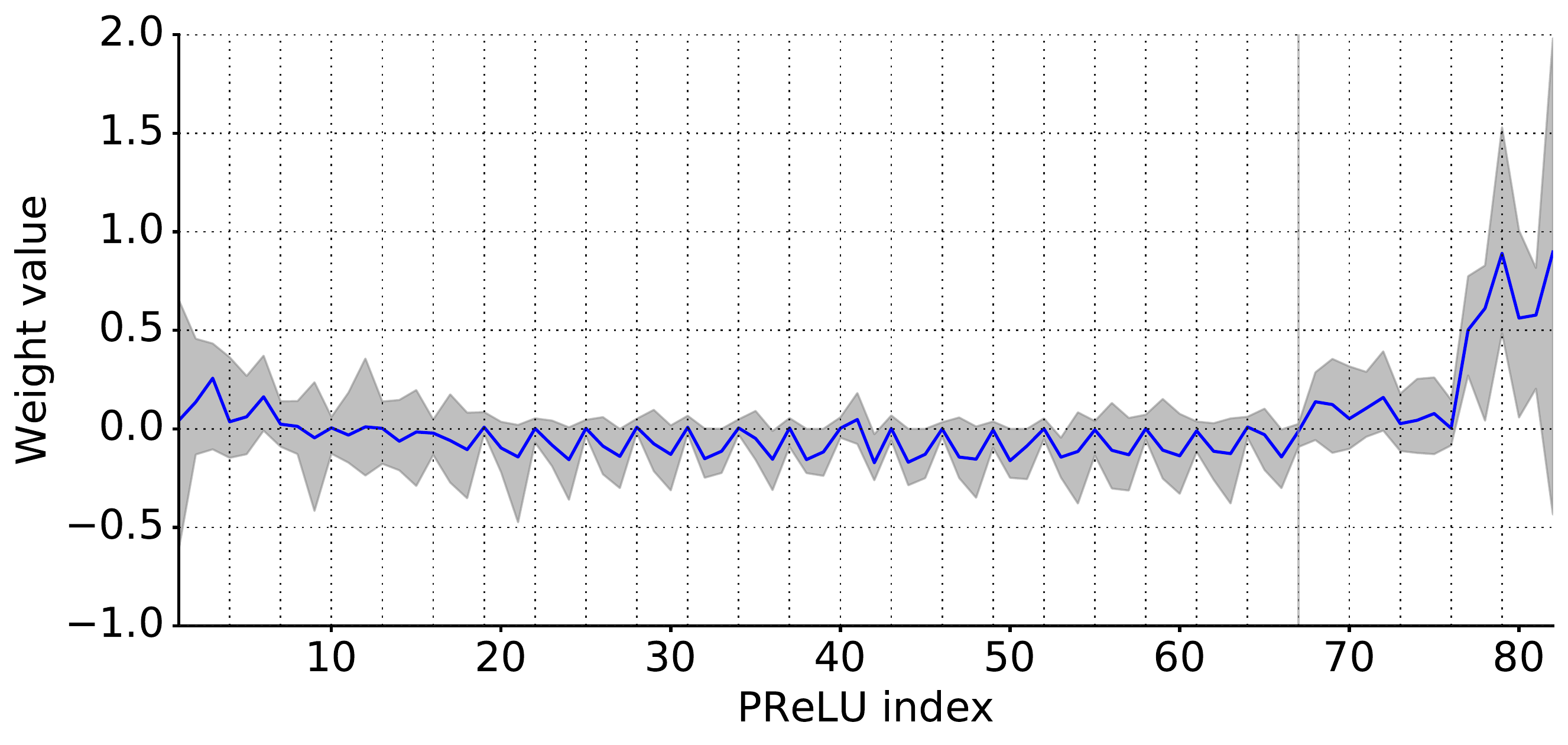}
  \caption{
    PReLU weight distribution vs network depth.
    Blue line is the weights mean, while an area between maximum and minimum weight is grayed out.
    Each vertical dotted line corresponds to a PReLU in the main branch and marks the boundary between each of bottleneck blocks.
    The gray vertical line at 67th module is placed at encoder-decoder border.
  }
  \vspace{0.05in}
  \label{fig:PReLU}
\end{figure}

We replaced all ReLUs in the network with PReLUs \cite{he2015}, which use an additional parameter per feature map, with the goal of learning the negative slope of non-linearities.
We expected that in layers where identity is a preferable transfer function, PReLU weights will have values close to $1$, and conversely, values around $0$ if ReLU is preferable.
Results of this experiment can be seen in Figure \ref{fig:PReLU}.

Initial layers weights exhibit a large variance and are slightly biased towards positive values, while in the later portions of the encoder they settle to a recurring pattern.
All layers in the main branch behave nearly exactly like regular ReLUs, while the weights inside bottleneck modules are negative i.e. the function inverts and scales down negative values.
We hypothesize that identity did not work well in our architecture because of its limited depth.
The reason why such lossy functions are learned might be that that the original ResNets \cite{he2016identity} are networks that can be hundreds of layers deep, while our network uses only a couple of layers, and it needs to quickly filter out information.
It is notable that the decoder weights become much more positive and learn functions closer to identity.
This confirms our intuitions that the decoder is used only to fine-tune the upsampled output.

\paragraph{Information-preserving dimensionality changes}
As stated earlier, it is necessary to downsample the input early, but aggressive dimensionality reduction can also hinder the information flow.
A very good approach to this problem has been presented in \cite{szegedy2015rethinking}.
It has been argued that a method used by the VGG architectures, i.e. as performing a pooling followed by a convolution expanding the dimensionality, however relatively cheap, introduces a representational bottleneck (or forces one to use a greater number of filters, which lowers computational efficiency).
On the other hand, pooling after a convolution, that increases feature map depth, is computationally expensive.
Therefore, as proposed in \cite{szegedy2015rethinking}, we chose to perform pooling operation in parallel with a convolution of stride 2, and concatenate resulting feature maps. 
This technique allowed us to speed up inference time of the initial block $10$ times.

Additionally, we have found one problem in the original ResNet architecture.
When downsampling, the first $1 \times 1$ projection of the convolutional branch is performed with a stride of $2$ in both dimensions, which effectively discards $75\%$ of the input.
Increasing the filter size to $2 \times 2$ allows to take the full input into consideration, and thus improves the information flow and accuracy.
Of course, it makes these layers $4 \times$ more computationally expensive, however there are so few of these in ENet, that the overhead is unnoticeable.

\paragraph{Factorizing filters}
It has been shown that convolutional weights have a fair amount of redundancy, and each $n \times n$ convolution can be decomposed into two smaller ones following each other: one with a $n \times 1$ filter and the other with a $1 \times n$ filter \cite{jin2014flattened}.
This idea has been also presented in \cite{szegedy2015rethinking}, and from now on we adopt their naming convention and will refer to these as asymmetric convolutions.
We have used asymmetric convolutions with $n = 5$ in our network, so cost of these two operations is similar to a single $3 \times 3$ convolution.
This allowed to increase the variety of functions learned by blocks and increase the receptive field.

What's more, a sequence of operations used in the bottleneck module (projection, convolution, projection) can be seen as decomposing one large convolutional layer into a series of smaller and simpler operations, that are its low-rank approximation.
Such factorization allows for large speedups, and greatly reduces the number of parameters, making them less redundant \cite{jin2014flattened}.
Additionally, it allows to make the functions they compute richer, thanks to the non-linear operations that are inserted between layers.

\paragraph{Dilated convolutions}
As argued above, it is very important for the network to have a wide receptive field, so it can perform classification by taking a wider context into account.
We wanted to avoid overly downsampling the feature maps, and decided to use dilated convolutions \cite{yu2015dilated} to improve our model.
They replaced the main convolutional layers inside several bottleneck modules in the stages that operate on the smallest resolutions.
These gave a significant accuracy boost, by raising IoU on Cityscapes by around $4$ percentage points, with no additional cost.
We obtained the best accuracy when we interleaved them with other bottleneck modules (both regular and asymmetric), instead of arranging them in sequence, as has been done in \cite{yu2015dilated}.

\paragraph{Regularization}
Most pixel-wise segmentation datasets are relatively small (on order of $10^3$ images), so such expressive models as neural networks quickly begin to overfit them.
In initial experiments, we used L2 weight decay with little success.
Then, inspired by \cite{huang2016stochastic}, we have tried stochastic depth, which increased accuracy.
However it became apparent that dropping whole branches (i.e. setting their output to $0$) is in fact a special case of applying Spatial Dropout \cite{tompson15}, where either all of the channels, or none of them are ignored, instead of selecting a random subset.
We placed Spatial Dropout at the end of convolutional branches, right before the addition, and it turned out to work much better than stochastic depth.

\section{Results} \label{results}

We benchmarked the performance of ENet on three different datasets to demonstrate real-time and accurate for practical applications.
We tested on CamVid and Cityscapes datasets of road scenes, and SUN RGB-D dataset of indoor scenes.
We set SegNet \cite{badrinarayanan15} as a baseline since it is one of the fastest segmentation models, that also has way fewer parameters and requires less memory to operate than FCN.
All our models, training, testing and performance evaluation scripts were using the Torch7 machine-learning library, with cuDNN backend.
To compare results, we use class average accuracy and intersection-over-union (IoU) metrics.

\subsection{Performance Analysis}

We report results on inference speed on widely used NVIDIA Titan X GPU as well as on NVIDIA TX1 embedded system module.
ENet was designed to achieve more than $10$ fps on the NVIDIA TX1 board with an input image size $640 \times 360$, which is adequate for practical road scene parsing applications.
For inference we merge batch normalization and dropout layers into the convolutional filters, to speed up all networks.

\newcommand{\resolution}[2]{\multicolumn{2}{c}{#1$\times$#2}}
\newcommand{\ms}{\multicolumn{1}{c}{ms}}
\newcommand{\fps}{\multicolumn{1}{c}{fps}}
\begin{table}[htb]
  \footnotesize
  \caption{Performance comparison.}
  \vspace{0.05in}
  \label{tab:speed}
  \centering
  \begin{tabular}{ l r r r r r r r r r r r r r r }
    \toprule
    \multirow{3}{*}{Model} &&\multicolumn{6}{c}{NVIDIA TX1} &  &\multicolumn{6}{c}{NVIDIA Titan X} \\
    \cmidrule{3-8} \cmidrule{10-15}
            &&\resolution{480}{320} &\resolution{640}{360}&\resolution{1280}{720} &
            &\resolution{640}{360}&\resolution{1280}{720}&\resolution{1920}{1080} \\
            &&\ms &\fps &\ms &\fps &\ms &\fps & &\ms&\fps &\ms&\fps &\ms&\fps \\
    \midrule
    SegNet  &&757    &1.3     &1251   &0.8   &-    &-    & &69     &14.6   &289  &3.5    &637    &1.6     \\
    ENet    &&47     &21.1    &69     &14.6  &262  &3.8  & &7      &135.4  &21   &46.8   &46     &21.6   \\
    \bottomrule
  \end{tabular}
\end{table}


\paragraph{Inference time} Table \ref{tab:speed} compares inference time for a single input frame of varying resolution. We also report the number of frames per second that can be processed.
Dashes indicate that we could not obtain a measurement, due to lack of memory.
ENet is significantly faster than SegNet, providing high frame rates for real-time applications and allowing for practical use of very deep neural network models with encoder-decoder architecture.

\begin{table}[htb]
  \caption{Hardware requirements. FLOPs are estimated for an input of $3\times640\times360$.}
  \vspace{0.05in}
  \label{tab:ops}
  \centering
  \begin{tabular}{ l r r r }
    \toprule
    &\multicolumn{1}{c}{GFLOPs} &\multicolumn{1}{c}{Parameters} & \multicolumn{1}{c}{Model size (fp16)}\\
    \midrule
    SegNet      &286.03                    &29.46M             & 56.2 MB\\
    ENet        &3.83                      &0.37M              & 0.7 MB\\
    \bottomrule
  \end{tabular}
\end{table}

\paragraph{Hardware requirements} Table \ref{tab:ops} reports a comparison of number of floating point operations and parameters used by different models.
ENet efficiency is evident, as its requirements are on two orders of magnitude smaller.
Please note that we report storage required to save model parameters in half precision floating point format.
ENet has so few parameters, that the required space is only 0.7MB, which makes it possible to fit the whole network in an extremely fast on-chip memory in embedded processors.
Also, this alleviates the need for model compression \cite{song15}, making it possible to use general purpose neural network libraries.
However, if one needs to operate under incredibly strict memory constraints, these techniques can still be applied to ENet as well.

\paragraph{Software limitations} One of the most important techniques that has allowed us to reach these levels of performance is convolutional layer factorization.
However, we have found one surprising drawback.
Although applying this method allowed us to greatly reduce the number of floating point operations and parameters, it also increased the number of individual kernels calls, making each of them smaller.

We have found that some of these operations can become so cheap, that the cost of GPU kernel launch starts to outweigh the cost of the actual computation.
Also, because kernels do not have access to values that have been kept in registers by previous ones, they have to load all data from global memory at launch, and save it when their work is finished.
This means that using a higher number of kernels, increases the number of memory transactions, because feature maps have to be constantly saved and reloaded.
This becomes especially apparent in case of non-linear operations.
In ENet, PReLUs consume more than a quarter of inference time.
Since they are only simple point-wise operations and are very easy to parallelize, we hypothesize it is caused by the aforementioned data movement.

These are serious limitations, however they could be resolved by performing kernel fusion in existing software i.e. create kernels that apply non-linearities to results of convolutions directly, or perform a number of smaller convolutions in one call.
This improvement in GPU libraries, such as cuDNN, could increase the speed and efficiency of our network even further.

\subsection{Benchmarks}

We have used the Adam optimization algorithm \cite{diederik14} to train the network.
It allowed ENet to converge very quickly and on every dataset we have used training took only 3-6 hours, using four Titan X GPUs.
It was performed in two stages: first we trained only the encoder to categorize downsampled regions of the input image, then we appended the decoder and trained the network to perform upsampling and pixel-wise classification.
Learning rate of $5\mathrm{e}{-4}$ and L2 weight decay of $2\mathrm{e}{-4}$, along with batch size of $10$ consistently provided the best results.
We have used a custom class weighing scheme defined as $w_{\mathrm{class}} = \frac{1}{\ln(c + p_{\mathrm{class}})}$.
In contrast to the inverse class probability weighing, the weights are bounded as the probability approaches $0$.
$c$ is an additional hyper-parameter, which we set to $1.02$ (i.e. we restrict the class weights to be in the interval of $\left[1, 50\right]$).

\begin{table}[htb]
  \small
  \caption{Cityscapes test set results}
  \vspace{0.05in}
  \label{tab:cityscape}
  \centering
  \begin{tabular}{ c c c c c }
    \toprule
    Model           &Class IoU      &Class iIoU     &Category IoU  &Category iIoU \\
    \midrule
    SegNet          &56.1           &34.2           &79.8          &\textbf{66.4} \\
    ENet            &\textbf{58.3}  &\textbf{34.4}  &\textbf{80.4} &64.0          \\
    \bottomrule
  \end{tabular}
\end{table}

\paragraph{Cityscapes}
This dataset consists of 5000 fine-annotated images, out of which 2975 are available for training, 500 for validation, and the remaining 1525 have been selected as test set \cite{cityscape2016}.
Cityscapes was the most important benchmark for us, because of its outstanding quality and highly varying road scenarios, often featuring many pedestrians and cyclists.
We trained on 19 classes that have been selected in the official evaluation scripts \cite{cityscape2016}.
It makes use of an additional metric called instance-level intersection over union metric (iIoU), which is IoU weighed by the average object size.
As reported in Table \ref{tab:cityscape}, ENet outperforms SegNet in class IoU and iIoU, as well as in category IoU.
ENet is currently the fastest model in the Cityscapes benchmark.
Example predictions for images from validation set are presented in Figure \ref{fig:cityscapes}.

\begin{table}[htb]
  \footnotesize
  \caption{Results on CamVid test set of (1) SegNet-Basic, (2) SegNet, and (3) ENet}
  \vspace{0.05in}
  \label{tab:camvid}
  \centering
  \begin{tabular}{ c c c c c c c c c c c c c c }
    \toprule
    \rotatebox[origin=c]{90}{Model}\hspace{0.07in} &\rotatebox[origin=c]{90}{Building} &\rotatebox[origin=c]{90}{Tree} &\rotatebox[origin=c]{90}{Sky} &\rotatebox[origin=c]{90}{Car} &\rotatebox[origin=c]{90}{Sign} &\rotatebox[origin=c]{90}{Road} &\rotatebox[origin=c]{90}{ Pedestrian } &\rotatebox[origin=c]{90}{Fence} &\rotatebox[origin=c]{90}{Pole} &\rotatebox[origin=c]{90}{Sidewalk} &\rotatebox[origin=c]{90}{Bicyclist} &\hspace{0.07in}\rotatebox[origin=c]{90}{Class avg.} &\rotatebox[origin=c]{90}{Class IoU} \\
    \midrule
    1\hspace{0.07in}   &75.0          &84.6           &91.2           &\textbf{82.7}  &36.9           &93.3           &55.0           &47.5           &\textbf{44.8}  &74.1           &16.0           &\hspace{0.07in}62.9 &n/a       \\
    2\hspace{0.07in}           &\textbf{88.8} &\textbf{87.3}  &92.4           &82.1           &20.5           &\textbf{97.2}  &57.1           &49.3           &27.5           &84.4           &30.7           &\hspace{0.07in}65.2       &\textbf{55.6}      \\
    3\hspace{0.07in}           &74.7          &77.8           &\textbf{95.1}  &82.4           &\textbf{51.0}  &95.1           &\textbf{67.2}  &\textbf{51.7}  &35.4           &\textbf{86.7}  &\textbf{34.1}  &\hspace{0.07in}\textbf{68.3}       &51.3      \\
    \bottomrule
  \end{tabular}
\end{table}

\paragraph{CamVid}

Another automotive dataset, on which we have tested ENet, was CamVid.
It contains 367 training and 233 testing images \cite{camvid08}.
There are eleven different classes such as building, tree, sky, car, road, etc. while the twelfth class contains unlabeled data, which we ignore while training.
The original frame resolution for this dataset is 960$\times$720 but we downsampled the images to 480$\times$360 before training.
In Table \ref{tab:camvid} we compare the performance of ENet with existing state-of-the-art algorithms.
ENet outperforms other models in six classes, which are difficult to learn because they correspond to smaller objects.
ENet output for example images from the test set can be found in Figure \ref{fig:camvid}.

\begin{table}[htb]
  \small
  \caption{SUN RGB-D test set results}
  \vspace{0.05in}
  \label{tab:sun1}
  \centering
  \begin{tabular}{ c c c c }
    \toprule
    Model           &Global avg.      &Class avg.     &Mean IoU    \\
    \midrule
    SegNet          &\textbf{70.3}    &\textbf{35.6}  &\textbf{26.3}\\
    ENet            &59.5             &32.6           &19.7         \\
    \bottomrule
  \end{tabular}

\end{table}

\paragraph{SUN RGB-D}
The SUN dataset consists of 5285 training images and 5050 testing images with 37 indoor object classes.
We did not make any use of depth information in this work and trained the network only on RGB data.
In Table \ref{tab:sun1} we compare the performance of ENet with SegNet \cite{badrinarayanan15}, which is the only neural network model that reports accuracy on this dataset.
Our results, though inferior in global average accuracy and IoU, are comparable in class average accuracy.
Since global average accuracy and IoU are metrics that favor correct classification of classes occupying large image patches, researchers generally emphasize the importance of other metrics in case of semantic segmentation.
One notable example is introduction of iIoU metric \cite{cityscape2016}.
Comparable result in class average accuracy indicates, that our network is capable of differentiating smaller objects nearly as well as SegNet.
Moreover, the difference in accuracy should not overshadow the huge performance gap between these two networks.
ENet can process the images in real-time, and is nearly $20 \times$ faster than SegNet on embedded platforms.
Example predictions from SUN test set are shown in Figure \ref{fig:sun}.

\begin{table}[!htb]
  \begin{adjustbox}{width=1.0\textwidth,center}
    \small
    \begin{tabular}{CMMMM}
      Input image
      &\includegraphics[width=.225\textwidth]{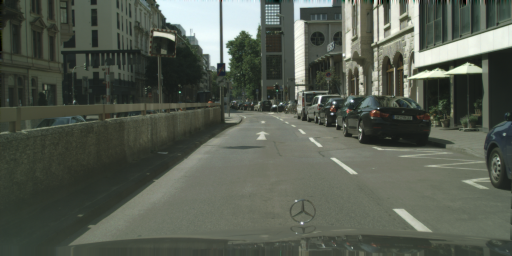}
      &\includegraphics[width=.225\textwidth]{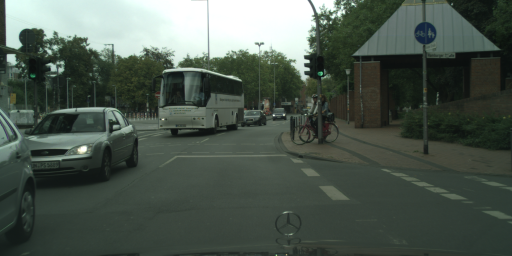}
      &\includegraphics[width=.225\textwidth]{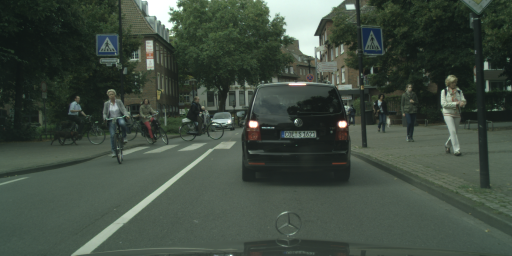}
      &\includegraphics[width=.225\textwidth]{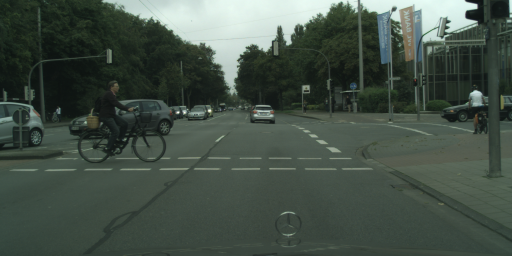}
      \\
      Ground truth
      &\includegraphics[width=.225\textwidth]{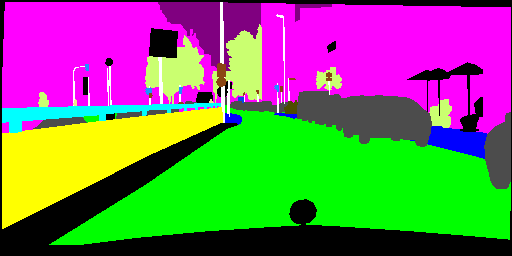}
      &\includegraphics[width=.225\textwidth]{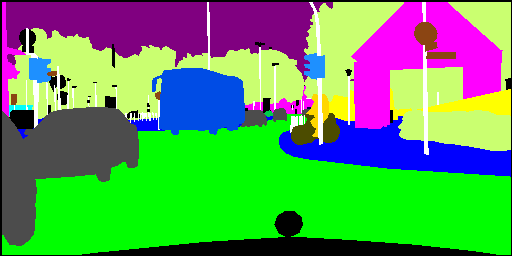}
      &\includegraphics[width=.225\textwidth]{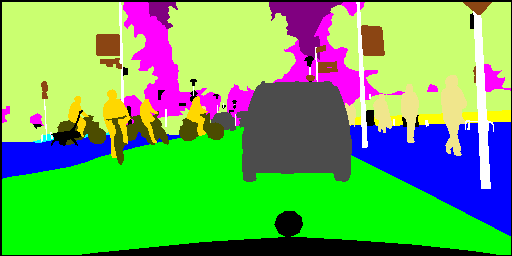}
      &\includegraphics[width=.225\textwidth]{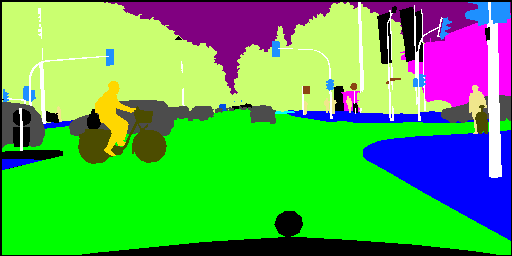}
      \\
      ENet output
      &\includegraphics[width=.225\textwidth]{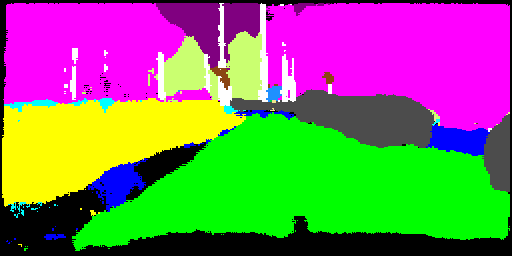}
      &\includegraphics[width=.225\textwidth]{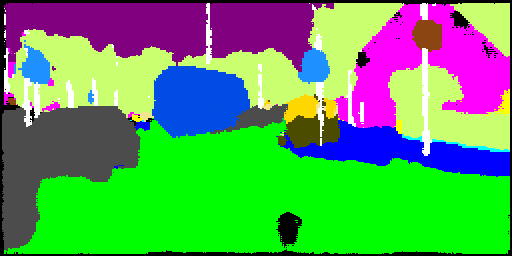}
      &\includegraphics[width=.225\textwidth]{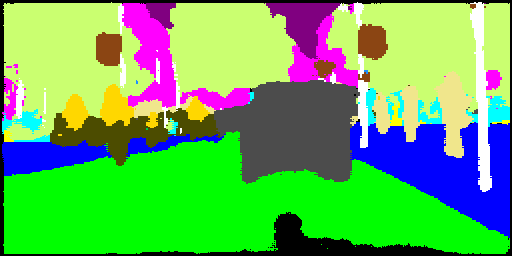}
      &\includegraphics[width=.225\textwidth]{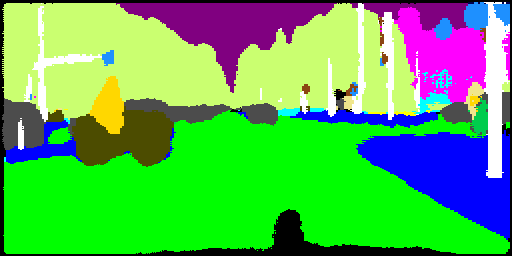}
      \\
    \end{tabular}
  \end{adjustbox}
  \vspace{0.05in}
  \captionof{figure}{ENet predictions on Cityscapes validation set \cite{cityscape2016}}
  \label{fig:cityscapes}

  \begin{adjustbox}{width=1.0\textwidth,center}
    \small
    \begin{tabular}{CMMMM}
      Input image
      &\includegraphics[width=.225\textwidth,trim={0 0 0 1cm},clip]{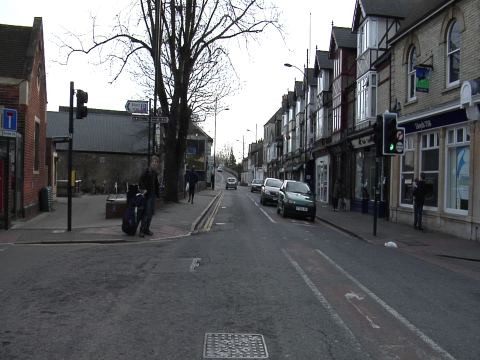}
      &\includegraphics[width=.225\textwidth,trim={0 0 0 1cm},clip]{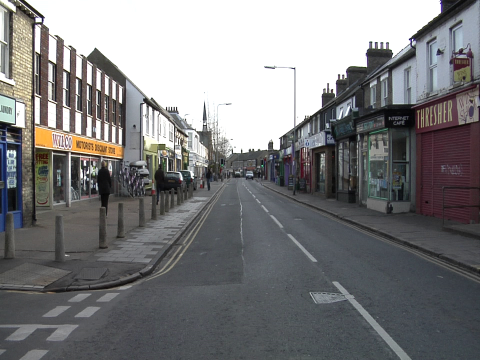}
      &\includegraphics[width=.225\textwidth,trim={0 0 0 1cm},clip]{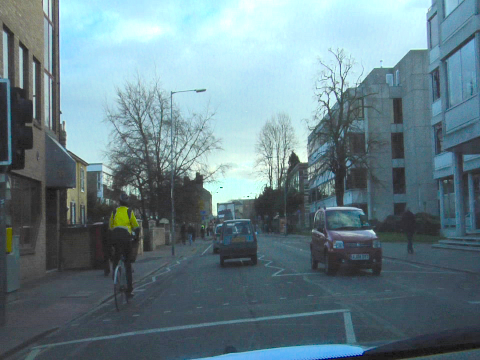}
      &\includegraphics[width=.225\textwidth,trim={0 0 0 1cm},clip]{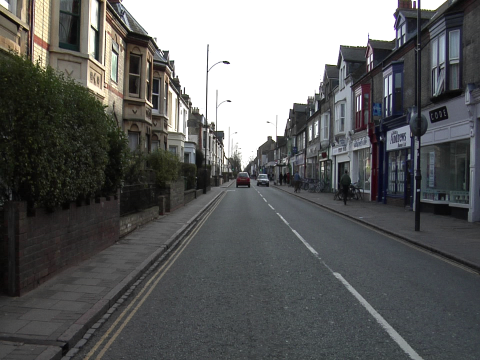}
      \\
      Ground truth
      &\includegraphics[width=.225\textwidth,trim={0 0 0 1cm},clip]{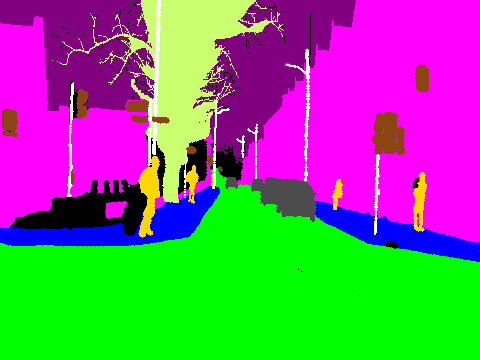}
      &\includegraphics[width=.225\textwidth,trim={0 0 0 1cm},clip]{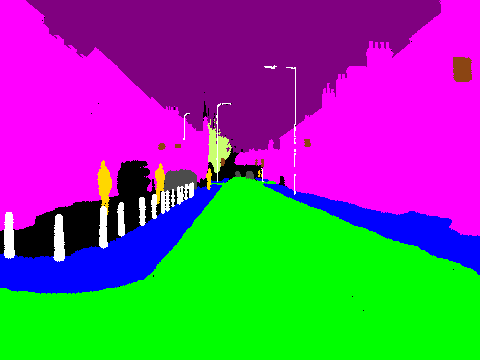}
      &\includegraphics[width=.225\textwidth,trim={0 0 0 1cm},clip]{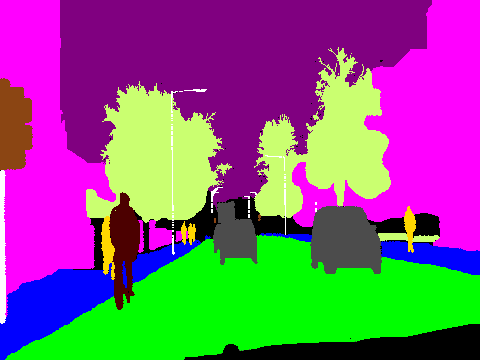}
      &\includegraphics[width=.225\textwidth,trim={0 0 0 1cm},clip]{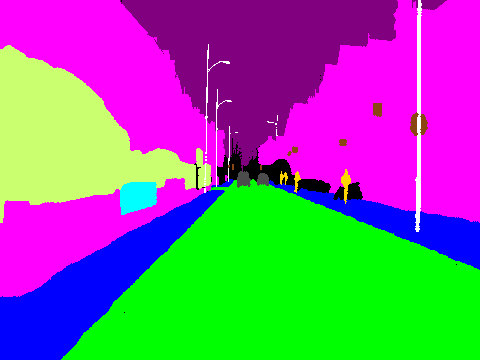}
      \\
      ENet output
      &\includegraphics[width=.225\textwidth,trim={0 0 0 1cm},clip]{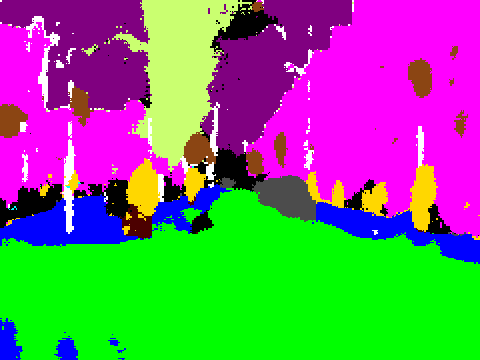}
      &\includegraphics[width=.225\textwidth,trim={0 0 0 1cm},clip]{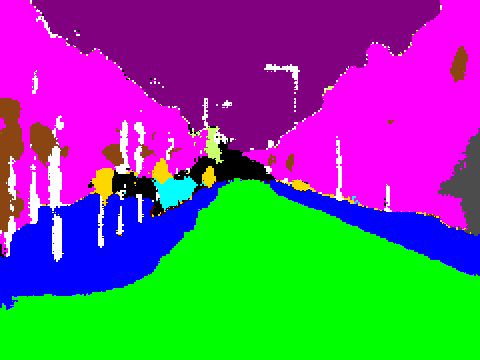}
      &\includegraphics[width=.225\textwidth,trim={0 0 0 1cm},clip]{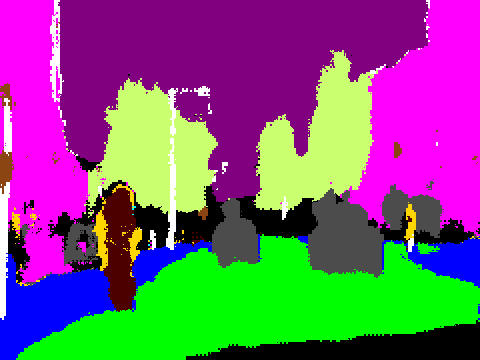}
      &\includegraphics[width=.225\textwidth,trim={0 0 0 1cm},clip]{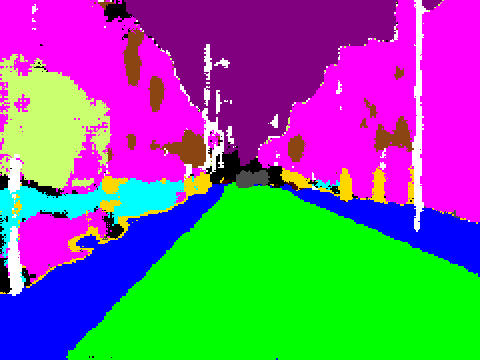}
      \\
    \end{tabular}
  \end{adjustbox}
  \vspace{0.05in}
  \captionof{figure}{ENet predictions on CamVid test set \cite{camvid08}}
  \label{fig:camvid}

  \begin{adjustbox}{width=1.0\textwidth,center}
    \small
    \begin{tabular}{CMMMM}
      Input image
      &\includegraphics[width=.225\textwidth,trim={0 0 0 1cm},clip]{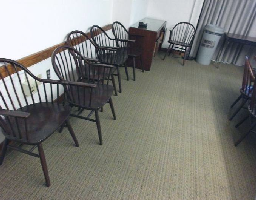}
      &\includegraphics[width=.225\textwidth,trim={0 0 0 1cm},clip]{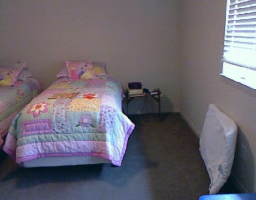}
      &\includegraphics[width=.225\textwidth,trim={0 0 0 1cm},clip]{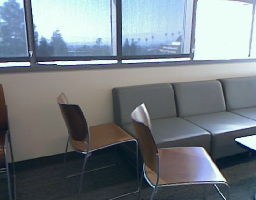}
      &\includegraphics[width=.225\textwidth,trim={0 0 0 1cm},clip]{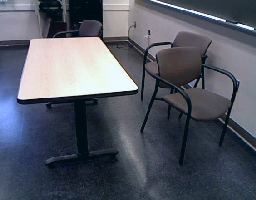}
      \\
      Ground truth
      &\includegraphics[width=.225\textwidth,trim={0 0 0 1cm},clip]{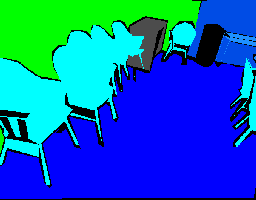}
      &\includegraphics[width=.225\textwidth,trim={0 0 0 1cm},clip]{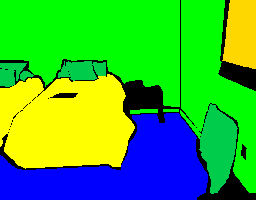}
      &\includegraphics[width=.225\textwidth,trim={0 0 0 1cm},clip]{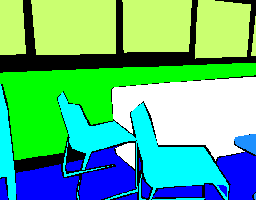}
      &\includegraphics[width=.225\textwidth,trim={0 0 0 1cm},clip]{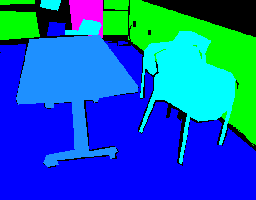}
      \\
      ENet output
      &\includegraphics[width=.225\textwidth,trim={0 0 0 1cm},clip]{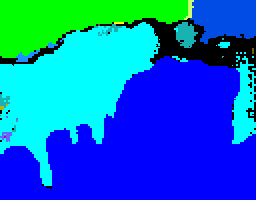}
      &\includegraphics[width=.225\textwidth,trim={0 0 0 1cm},clip]{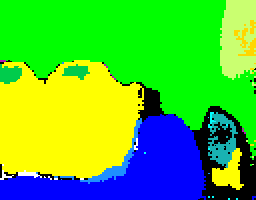}
      &\includegraphics[width=.225\textwidth,trim={0 0 0 1cm},clip]{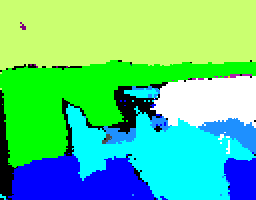}
      &\includegraphics[width=.225\textwidth,trim={0 0 0 1cm},clip]{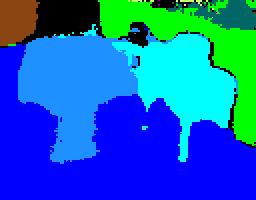}
      \\
    \end{tabular}
  \end{adjustbox}
  \vspace{0.05in}
  \captionof{figure}{ENet predictions on SUN RGB-D test set \cite{sun2015}}
  \label{fig:sun}
\end{table}

\section{Conclusion} \label{conclusion}

We have proposed a novel neural network architecture designed from the ground up specifically for semantic segmentation.
Our main aim is to make efficient use of scarce resources available on embedded platforms, compared to fully fledged deep learning workstations.
Our work provides large gains in this task, while matching and at times exceeding existing baseline models, that have an order of magnitude larger computational and memory requirements.
The application of ENet on the NVIDIA TX1 hardware exemplifies real-time portable embedded solutions.

Even though the main goal was to run the network on mobile devices, we have found that it is also very efficient on high end GPUs like NVIDIA Titan X.
This may prove useful in data-center applications, where there is a need of processing large numbers of high resolution images.
ENet allows to perform large-scale computations in a much faster and more efficient manner, which might lead to significant savings.

\section*{Acknowledgment}

This work is partly supported by the Office of Naval Research (ONR) grants N00014-12-1-0167, N00014-15-1-2791 and MURI N00014-10-1-0278.
We gratefully acknowledge the support of NVIDIA Corporation with the donation of the TX1, Titan X, K40 GPUs used for this research.

\pagebreak

\bibliographystyle{IEEEtran}
{
\fontsize{9}{9}
\selectfont
\setlength{\bibsep}{0pt}
\bibliography{ref}
}

\end{document}